\pdfoutput=1
\documentclass[11pt,a4paper]{article}
\usepackage[hyperref]{emnlp2020}
\usepackage{times}
\usepackage{latexsym}
\usepackage{multirow, makecell}

\usepackage{microtype}
\usepackage{amsmath}
\usepackage{graphicx}
\usepackage{todonotes}
\usepackage{enumitem}
\usepackage{svg}
\usepackage{bm}
\usepackage{hyperref}

\aclfinalcopy 

\newcommand{\GAENet}{\textrm{GEANet}}
\newcommand{\GE}{\textrm{{\fontfamily{qcr}\selectfont GE'11} }}
\newcommand{\GAENetFull}{Graph Edge-conditioned Attention Networks}

\makeatletter
\renewcommand\paragraph{\@startsection{paragraph}{4}{\z@}%
                                    {.25ex \@plus.5ex \@minus.2ex}%
                                    {-1em}%
                                    {\normalfont\normalsize\bfseries}}

\makeatother

\title{Biomedical Event Extraction with Hierarchical Knowledge Graphs} 

\author{
    Kung-Hsiang Huang \textsuperscript{\rm 1}~~~
    Mu Yang\textsuperscript{\rm 1}~~~
    Nanyun Peng\textsuperscript{\rm 1,2}\\
    \textsuperscript{\rm 1} Information Sciences Institute, University of Southern California\\
    \textsuperscript{\rm 2} Computer Science Department, University of California, Los Angeles \\
    {\tt \{kunghsia, yangmu\}@usc.edu}\\
    {\tt violetpeng@cs.ucla.edu} 
}
\date{}

\begin{document}
\maketitle

\begin{abstract}

Biomedical event extraction is critical in understanding biomolecular interactions described in scientific corpus. One of the main challenges is to identify nested structured events that are associated with non-indicative trigger words. We propose to incorporate domain knowledge from Unified Medical Language System (UMLS) to a pre-trained language model via a hierarchical graph representation encoded by a proposed \GAENetFull{} (\GAENet). To better recognize the trigger words, each sentence is first grounded to a sentence graph based on a jointly modeled hierarchical knowledge graph from UMLS. The grounded graphs are then propagated by \GAENet, a novel graph neural networks for enhanced capabilities in inferring complex events. On BioNLP 2011 GENIA Event Extraction task, our approach achieved 1.41\% $F_{1}$ and 3.19\% $F_{1}$ improvements on all events and complex events, respectively. Ablation studies confirm the importance of \GAENet{} and hierarchical KG. 

\end{abstract}
\section{Introduction}


Biomedical event extraction is a task that identifies a set of actions among proteins or genes that are associated with biological processes from natural language texts~\cite{kim-etal-2009-overview, kim2011overview}. Development of biomedical event extraction tools enables many downstream applications, such as domain-specific text mining \cite{ananiadou2015event,spangher2020enabling}, semantic search engines~\cite{miyao-etal-2006-semantic} and automatic population and enrichment of database ~\cite{hirschman2012text}. 

A typical event extraction system 1) finds triggers that most clearly demonstrate the presence of events, 2) recognizes the protein participants (arguments), and 3) associates the arguments with the corresponding event triggers. For instance, the sentence ``Protein A inhibits the expression of Protein B" will be annotated with two nested events: \textit{Gene expression}(Trigger: expression, Arg-Theme: 
Protein B) and \textit{Negative Regulation}(Trigger: inhibits, Arg-Theme: \textit{Gene expression}(Protein B), Arg-Cause: Protein A).

\begin{figure}
  \centering
  \includegraphics[width=.9\linewidth]{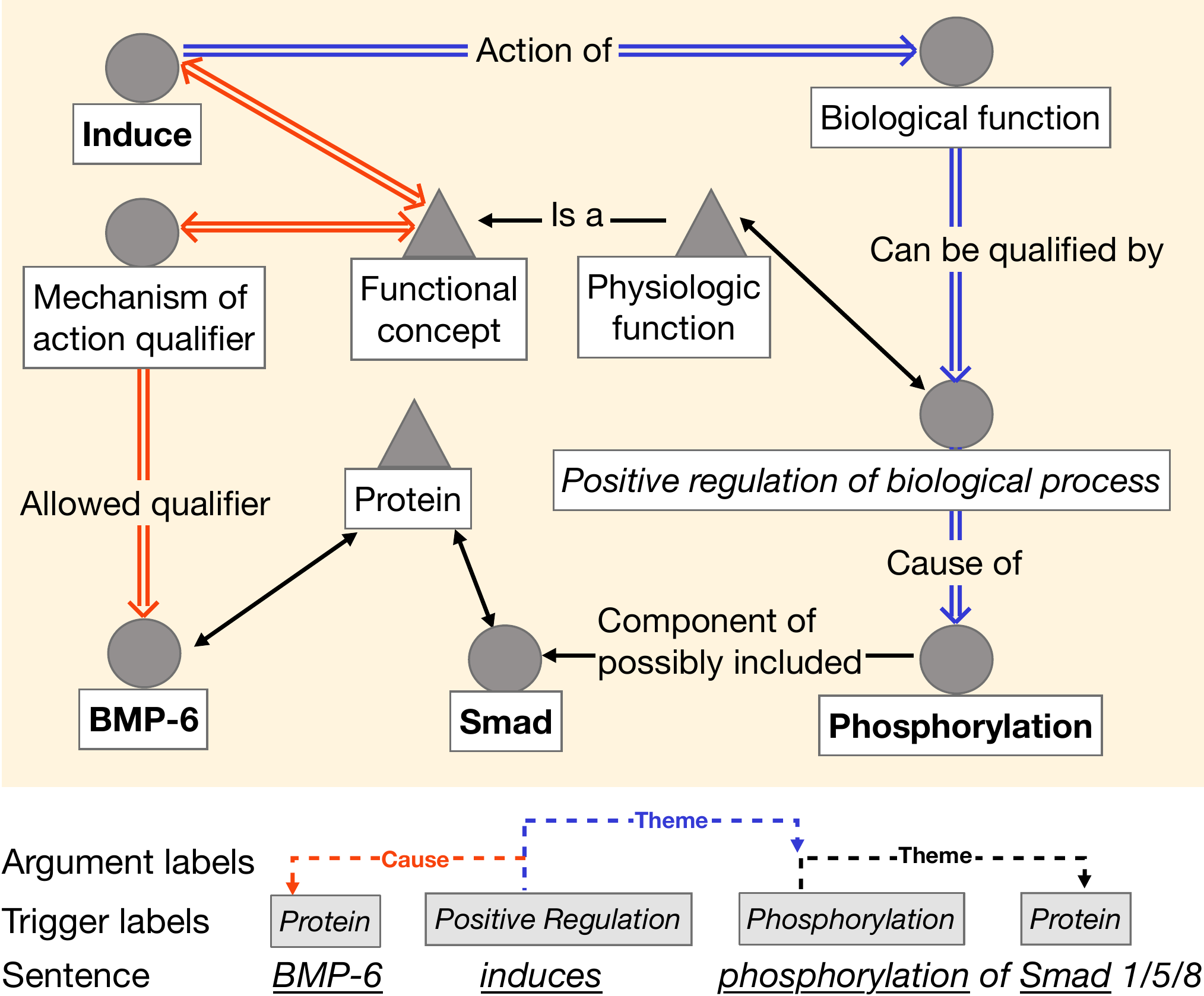}
  \caption{An example of a UMLS-based hierarchical KG assisting event extraction. Circles represent concept nodes and triangles represent semantic nodes. Nodes associated with the tokens in the example sentence are boldfaced. Bidirectional edges imply hierarchical relation between concept and semantic nodes.  The word ``induces" is a trigger of a \textit{Positive regulation} event, whose trigger role and corresponding argument role cannot be easily determined with only textual input. The KG provides clues for identifying this trigger and its corresponding arguments given the red and blue double line reasoning paths connecting nodes \textbf{BMP-6}, \textbf{Induce}, \textbf{Phosphorylation}, and \textit{Positive regulation of biological process}. We can infer that: 1) ``induces" is an action of biological function, 2) a biological function can be quantified by positive regulation, and 3) positive regulation can result in phosphorylation.}
  \label{toy_example}
  \vspace{-1.5em}
\end{figure} 

Early attempts on biomedical event extraction adopted hand-crafted features \cite{bjorne-etal-2009-extracting, bjorne-salakoski-2011-generalizing, riedel2011fast, venugopal2014relieving}. Recent advances have shown improvements using deep neural networks via distributional word representations in the biomedical domain \cite{moen2013distributional, rao2017biomedical, bjorne-salakoski-2018-biomedical, shafieibavani2019global}. \newcite{li2019biomedical} further extends the word representations with embeddings of descriptive annotations from a knowledge base and demonstrates the importance of domain knowledge in biomedical event extraction. 

However, encoding knowledge with distributional embeddings does not provide adequate clues for identifying challenging events with non-indicative trigger words and nested structures. These embeddings do not contain \textit{structural} or \textit{relational} information about the biomedical entities. To overcome this challenge, we present a framework that incorporates knowledge from hierarchical knowledge graphs with graph neural networks (GNN) on top of a pre-trained language model.

\textbf{Our first contribution} is a novel representation of knowledge as hierarchical knowledge graphs containing both \textit{conceptual} and \textit{semantic} reasoning paths that enable better trigger and word identification based on Unified Medical Language System (UMLS), a biomedical knowledge base. Fig. \ref{toy_example} shows an example where the \textit{Positive Regulation} event can be better identified with knowledge graphs and factual relational reasoning. \textbf{Our second contribution} is a new GNN, \GAENetFull{} (\GAENet), that encodes complex domain knowledge. By integrating edge information into the attention mechanism, \GAENet{} has greater capabilities in reasoning the plausibility of different event structure through factual relational paths in knowledge graphs (KGs).

Experiments show that 
our proposed method achieved state-of-the-art results on the BioNLP 2011 event extraction task~\cite{kim2011overview}.\footnote{Our code for pre-proecessing, modeling,  and evaluation is available at \url{https://github.com/PlusLabNLP/GEANet-BioMed-Event-Extraction}.}

\section{Background} \label{sec:background}

\paragraph{UMLS Knowledge Base.}
\label{UMLS_KB}
Unified Medical Language System (UMLS) is a knowledge base for biomedical terminology and standards, which includes three knowledge sources: \textit{Metathesaurus}, \textit{Semantic Network}, and \textit{Specialist Lexicon and Lexical Tools}~\cite{bodenreider2004unified}. We use the former two sources to build hierarchical KGs. The \textit{concept network} from Metathesaurus contains the relationship between each biomedical concept pairs, while each concept contains one or more semantic types that can be found in the \textit{semantic network}. 
The \textit{concept network} provides direct definition lookup of the recognized biomedical terms, while the \textit{semantic network} supports with additional knowledge in the semantic aspect. Example tuples can be found in Figure \ref{toy_example}.\footnote{There are several bi-directional relations between some concepts. We only show one direction for simplicity.} There are 3.35M concepts, 10 concept relations, 182 semantic types, and 49 semantic relations in total. 

\section{Proposed Approach}


Our event extraction framework builds upon the pre-trained language model, SciBERT \cite{beltagy-etal-2019-scibert}, and supplement it with a novel graph neural network model, \GAENet{}, that encodes domain knowledge from hierarchical KGs. We will first illustrate each component and discuss how training and inference are done.  

\begin{figure}
  \centering
  \includegraphics[width=\linewidth]{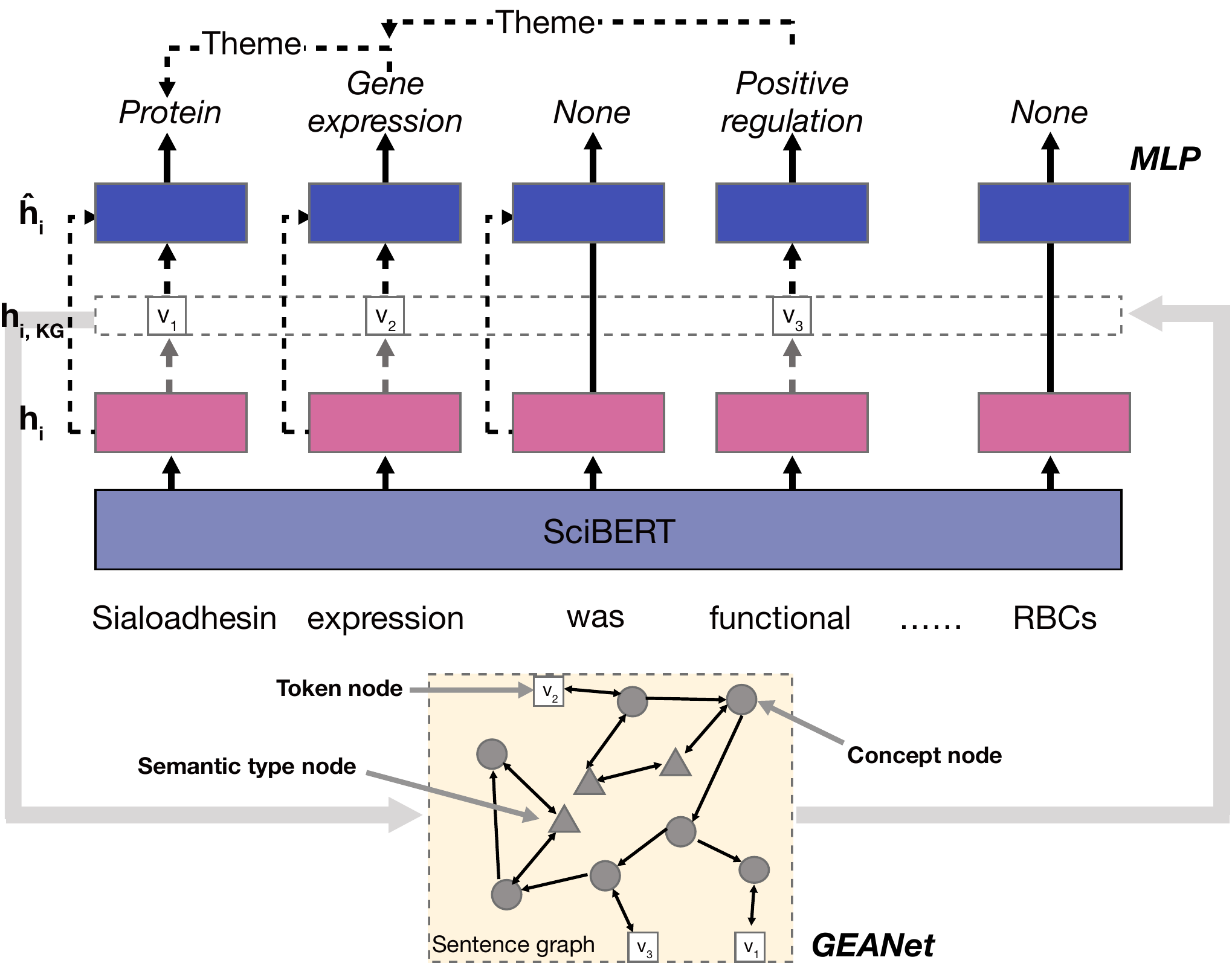}
  \caption{Overview of knowledge incorporation. Contextualized embeddings for each token are generated by SciBERT. \GAENet{} updates node embeddings for $v_{1}, v_{2}, \textrm{and } v_{3}$ via corresponding sentence graph.}
  \label{KI_figure}
\vspace{-1em}
\end{figure}


\subsection{Hierarchical Knowledge Graph Modeling}
\label{sec:hierarchical_kg}
The two knowledge sources discussed in Section \ref{UMLS_KB} are jointly modeled as a \textit{hierarchical graph} for each sentence, which we refer to as a \textit{sentence graph}. Each sentence graph construction consists of three steps: concept mapping, concept network construction, and semantic type augmentation.

The first step is to map each sentence in the corpus to \textit{UMLS biomedical concepts} with MetaMap, an entity mapping tool for UMLS concepts \cite{aronson2001effective}. 
There are 7903 concepts (entities) being mapped from the corpus, denoted as \textit{K}. 
The next step is \textit{concept network construction}, where a minimum spanning tree (MST) that connects mapped concepts in the previous step is identified, forming concept reasoning paths. 
This step is NP-complete.\footnotemark\footnotetext{Finding a MST on a subset of nodes (\textit{K}) is known as a Steiner tree problem.} 
We adopt a 2-approximate solution that constructs a global MST for the corpora \GE{} by running breadth-first search, assuming all edges are of unit distance. To prune out less relevant nodes and to improve computation efficiency, concept nodes that are not in \textit{K} with less than \textit{T} neighbors in \textit{K} are removed.\footnotemark
\footnotetext{\textit{T} is empirically set to be 35.}
The spanning tree for each sentence is then obtained by depth-first search on the global MST.
Each matched token in the corpus is also included as a token node in the sentence graph, connecting with corresponding concept node. 
Finally, the semantic types for each concept node are modeled as nodes that are linked with associated concept nodes in the sentence graph. Two semantic type nodes will also be linked if they have known relationships in the \textit{semantic network}. 

\subsection{\GAENet}
The majority of existing graph neural networks (GNN) consider only hidden states of nodes and adjacency matrix without modeling edge information. 
To properly model the hierarchy of the graph, it is essential for the message passing function of a GNN to consider edge features. We propose Graph Edge Conditioned Attention Networks (\GAENet) to integrate edge features into the attention mechanism for message propagation. 
The node embeddings update of \GAENet{} at the $l$-th layer can be expressed as follows:

{\small
\begin{align}
  \bm{x}^{(l)}_{i} &= \textrm{MLP}_{\theta} \bm{x}^{(l-1)}_{i} + \sum_{j \in \mathcal{N}(i)} a_{i,j} \cdot \bm{x}^{(l-1)}_{j}\\
  a_{i,j} &=
        \frac{
        \exp\left( \textrm{MLP}_{\psi}( \bm{e}_{i,j} )
        \right)}
        {\sum_{k \in \mathcal{N}(i) }
        \exp\left( \textrm{MLP}_{\psi}( \bm{e}_{i,k} )
        \right)}
\end{align}}

where $\bm{x}^{(l)}_{i}$ denotes the node embeddings at layer $l$, $\bm{e}_{i,j}$ denotes the embedding for edge $(i, j)$, and $\textrm{MLP}_{\psi}$ and $\textrm{MLP}_{\theta}$ are two multi-layer perceptrons.

\GAENet{} is inspired by Edge Conditioned Convolution (ECC), where convolution operation depends on edge type \cite{simonovsky2017dynamic},

{\small
\begin{equation}
  \bm{x}^{(l)}_{i} = \textrm{MLP}_{\theta} \bm{x}^{(l-1)}_{i} + \sum_{j \in \mathcal{N}(i)}  \bm{x}^{(l-1)}_{j} \cdot \textrm{MLP}_{\psi}(\bm{e}_{i,j})
\end{equation}
}
Compared to ECC, \GAENet{} is able to determine the relative importance of neighboring nodes with attention mechanism.


\paragraph{Knowledge Incorporation.}
\label{knowledge_incorporation}
We build \GAENet{} on top of SciBERT~\cite{peters-etal-2019-knowledge} 
to incorporate domain knowledge into rich contextualized representations. Specifically, we take the contextual embeddings $\{\bm{h}_{1},..., \bm{h}_{n}\}$ produced by SciBERT as inputs and produces knowledge-aware embeddings $\{\bm{\hat{h}}_{1},..., \bm{\hat{h}}_{n}\}$ as outputs. 
To initialize the embeddings for a sentence graph, for a mapped token, we project its SciBERT contextual embedding to initialize its corresponding node embedding $\bm{h}_{i, \textrm{KG}} = \bm{h}_{i}\bm{W}_{\textrm{KG}} + \bm{b}_{\textrm{KG}}$. Other nodes and edges are initialized by pretrained KG embeddings (details in Section~\ref{sec:setup}). To accommodate multiple relations between two entities in UMLS, edge embeddings $\bm{e}_{i,j}$ are initialized by summing the embeddings of each relation between the nodes $i$ and $j$. Then we apply layers of \GAENet{} to encode the graph $\bm{h}^{l}_{i, \textrm{KG}} = \textrm{\GAENet}(\bm{h}_{i, \textrm{KG}}) $.  The knowledge-aware representation is obtained by aggregating SciBERT representations and KG representations, $\bm{\hat{h}}_{i}  = \bm{h}^{l}_{i, \textrm{KG}}\bm{W}_{\textrm{LM}} + \bm{b}_{\textrm{LM}} + \bm{h}_{i}$.\footnotemark ~The process is illustrated in Figure \ref{KI_figure} \GAENet{} layer.
\footnotetext{$\bm{\hat{h}}_{i} = \bm{h}_{i}$ for each token \textit{i} without mapped concept.}


\subsection{Event Extraction}
The entire framework is trained with a multitask learning pipeline consisting of trigger classification and argument classification, following \cite{han2019deep,han-etal-2019-joint}. Trigger classification predicts the trigger type for each token. 
The predicted score of each token is computed as $\bm{\hat{y}}^{tri}_{i} = \textrm{MLP}^{tri}(\bm{\hat{h}}_{i})$. In the argument classification stage, each possible pair of gold trigger and gold entity is gathered and labeled with corresponding argument role.\footnotemark\footnotetext{During inference, predicted triggers are used instead.} The argument scores between the $i$-th token and $j$-th token are computed as $\bm{\hat{y}}^{arg}_{i,j} = \textrm{MLP}^{arg}(\hat{\bm{h}_{i}};\hat{\bm{h}_{j}})$, where $(;)$ denotes concatenation. Cross Entropy loss $\mathcal{L}^{t} = - \frac{1}{N^{t}}\sum_{i=1}^{N^{t}}  \bm{y}^{t}_{i} \cdot \log\bm{\hat{y}}^{t}_{i}, \label{eq:5}$  is used for both tasks, 
%
where $t$ denotes task, $N^{t}$ denotes the number of training instances of task $t$, $\bm{y}^{t}_{i}$ denotes the ground truth label, and $\bm{\hat{y}}^{t}_{i}$ denotes the predicted label. The multitask learning minimizes the sum of the two losses $\mathcal{L} = \mathcal{L}^{tri} + \mathcal{L}^{arg}$ in the training stage. During inference, unmerging is conducted to combine identified triggers and arguments for multiple arguments events \cite{bjorne-salakoski-2011-generalizing}. We adopted similar unmerging heuristics. For \textit{Regulation} events, we use the same heuristics as \newcite{bjorne-etal-2009-extracting}. For \textit{Binding} events, we subsume all \textit{Theme} arguments associated with a trigger into one event such that every trigger corresponds to only one single \textit{Binding} event. 




\section{Experiments}

\begin{table}[t]
\small
\centering
\begin{tabular}{llccc}
\hline & \textbf{Model}  & \textbf{Recall}  & \textbf{Prec.}  & \textbf{F1} \\ \hline
\multirow{4}{*}{Prior} & TEES & 49.56 & 57.65 & 53.30 \\
& Stacked Gen. & 48.96 &  66.46 & 56.38 \\
& TEES CNN & 49.94 & 69.45 & 58.10        \\
& KB-driven T-LSTM & 52.14 & 67.01 & 58.65 \\
\hline
\hline
\multirow{2}{*}{Ours} & SciBERT-FT & 53.89 & 63.97 & 58.50 \\
& \GAENet-SciBERT & \textbf{56.11}  &  64.61 &  \textbf{60.06} \\
\hline
\end{tabular}
\caption{\label{font-table} Model comparison on \GE test set.}
\label{overall_comparison}
\end{table}

\begin{table}[t]
\small
\centering
\begin{tabular}{@{\ \ }l|c@{\ \ }c@{\ \ }c@{\ \ }}
\hline \textbf{Model} & \textbf{Recall}  & \textbf{Prec.}  & \textbf{F1} \\
\hline

 KB-driven T-LSTM  & 41.73 & 55.73  & 47.72 \\
\hline
\hline

 SciBERT-FT  & 45.39 & 54.48 & 49.52 \\
 \GAENet- SciBERT  & \textbf{47.23} &  55.21 & \textbf{50.91} \\
 
\hline
\end{tabular}
\caption{\label{font-table} Performance comparison on the \textit{Regulation} events of the test set (including \textit{Regulation, Positive Regulation}, and \textit{Negative Regulation} sub-events). }
\label{regulation_comparison}
\vspace{-1em}
\end{table}

\subsection{Experimental Setup} \label{sec:setup}
Our models are evaluated on BioNLP 11 GENIA event extraction task (\GE). All models were trained on the training set, validated on the dev set, and tested on the test set. A separate evaluation on \textit{Regulation} events is conducted to validate the effectiveness of our framework on nested events with non-indicative trigger word. Reported results are obtained from the official evaluator under approximate span and recursive criteria.

In the preprocessing step, the \GE{} corpora were parsed with TEES preprocessing pipeline \cite{bjorne-salakoski-2018-biomedical}. Tokenization is done by the SciBERT tokenizer. Biomedical concepts in each sentence are then recognized with MetaMap and aligned with their corresponding tokens. The best performing model was found by grid search conducted on the dev set. The edge and node representation in KGs were intialized with 300 dimensional pre-trained embeddings using TransE \cite{wang2014knowledge}. The entire framework is optimized with BERTAdam optimizer for a maximum of 100 epochs with batch size of 4. Training is stopped if the dev set $F_{1}$ does not improve for 5 consecutive epochs (more details see Appendix). 

\begin{table}[t]
\small
\centering
\begin{tabular}[ht!]{lcc} 
\hline \textbf{Model} & \textbf{Dev F1} & \textbf{Test F1} \\ \hline
\GAENet-SciBERT & \textbf{60.38} & \textbf{60.06} \\
\quad - \GAENet & 59.33 & 58.50 \\
\quad - STY nodes & 60.12 & 59.34 \\
\quad \GAENet $\rightarrow$ ECC & 58.50 & 58.27 \\
\quad \GAENet $\rightarrow$ GAT & 59.55 & 59.87 \\

\hline
\end{tabular}
\caption{Ablation study over different components. \label{ablation}}
\vspace{-1em}
\end{table}
\subsection{Results and Analysis}
\paragraph{Comparison with existing methods}
We compare our method with the following prior works: \textbf{TEES} and \textbf{Stacked Gen.} use SVM-based models with token and sentence-level features \cite{bjorne-salakoski-2011-generalizing, majumder-etal-2016-biomolecular}; \textbf{TEES CNN} leverages Convolutional Neural Networks and dependency parsing graph \cite{bjorne-salakoski-2018-biomedical}; \textbf{KB-driven T-LSTM} adopts an external knowledge base with type and sentence embeddings, into a Tree-LSTM model \cite{li2019biomedical}. \textbf{SciBERT-FT} is a fine-tuned SciBERT without external resources, the knowledge-agnostic counterpart of \textbf{\GAENet-SciBERT}. According to Table \ref{overall_comparison}, \textbf{SciBERT-FT} achieves similar performance to \textbf{KB-driven T-LSTM}, implying that SciBERT may have stored domain knowledge implicitly during pre-training. Similar hypothesis has also been studied in commonsense reasoning \cite{wang-etal-2019-make}. \textbf{\GAENet-SciBERT} achieves an absolute improvement of 1.41\%  in $F_{1}$ on the test data compared to the previous state-of-the-art method. In terms of \textit{Regulation} events, Table \ref{regulation_comparison} shows that \textbf{\GAENet-SciBERT} outperforms the previous system and fine-tuned SciBERT by 3.19\% and 1.39\% in $F{1}$. 

\paragraph{Ablation study}
To better understand the importance of different model components, ablation study is conducted and summarized in Table \ref{ablation}. \GAENet{} achieves the highest $F_{1}$ when compared to two other GNN variants, ECC and GAT \cite{velickovic2018graph}, demonstrating its stronger knowledge incorporation capacity. Hierarchical knowledge graph representation is also shown to be critical. Removing semantic type (STY) nodes from hierarchical KGs leads to performance drop.

\begin{figure}
  
  \centering
  \includegraphics[width=\linewidth]{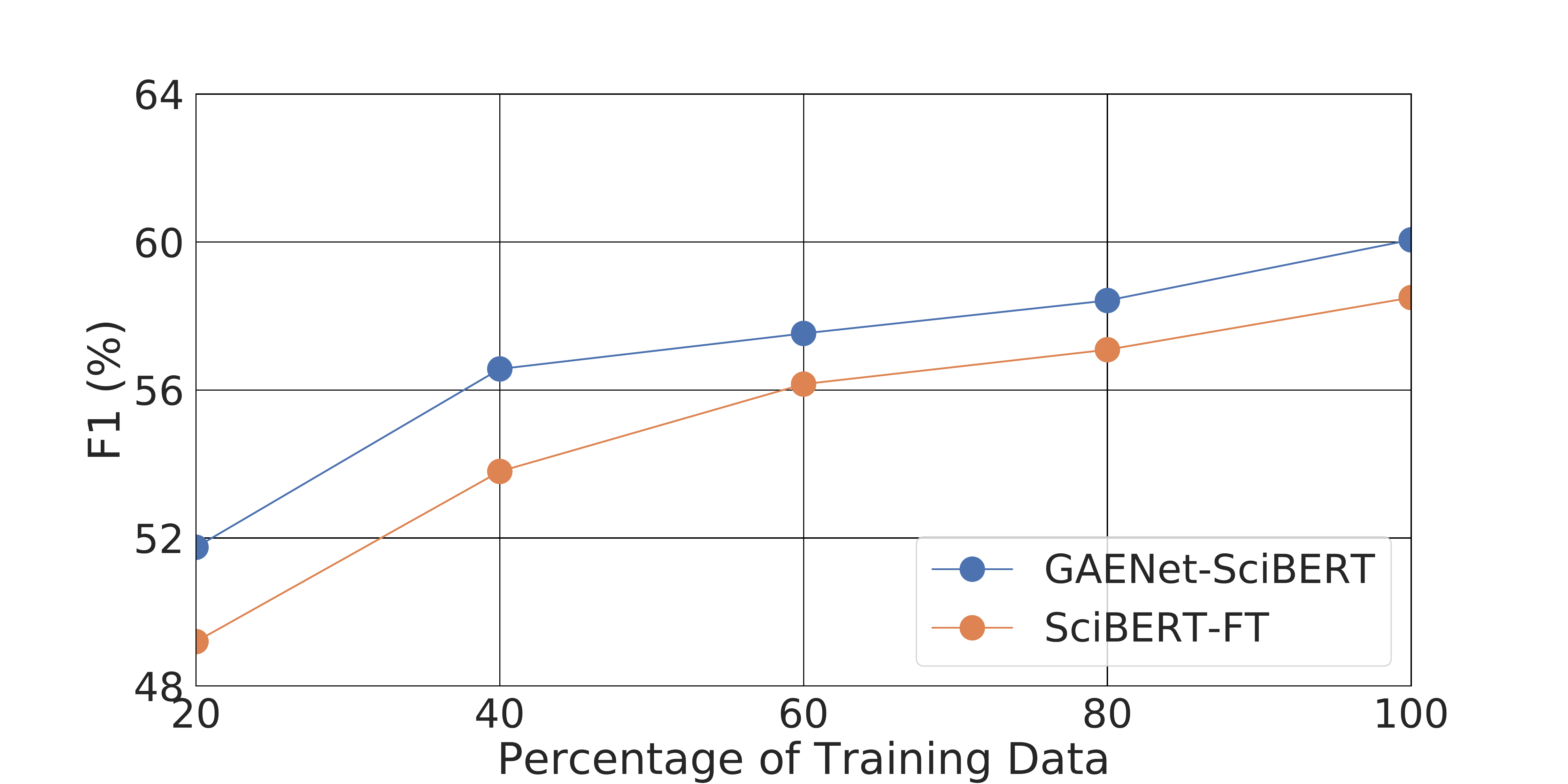}
  \caption{Performance comparison on the test set w.r.t. different amount of training data.}
  \label{label_efficiency}
  
\end{figure}

\paragraph{Impact of amount of training data} Model performance on different amount of randomly sampled training data is shown in Fig. \ref{label_efficiency}. \textbf{\GAENet-SciBERT} shows consistent improvement over fine-tuned SciBERT across different fractions. The performance gain is slightly larger with less training data. This illustrates the robustness of \GAENet{} in integrating domain knowledge and its particular advantage under low-resource setting.

\paragraph{Error Analysis}
By comparing the predictions from \textbf{\GAENet-SciBERT} and gold events in the dev set, two major failed cases are identified:

\begin{itemize}

  \item \textbf{Adjective Trigger}: Most events are associated with a verb or noun trigger. Adjective triggers are scarce in the training set ($\sim$7\%), which poses a challenge to identify this type of trigger. Although knowledge-aware methods should be able to resolve these errors theoretically, these adjective triggers often cannot be linked with UMLS concepts. Without proper grounding, it is hard for our model to recognize these triggers.
  \item \textbf{Misleading Trigger}: Triggers providing ``clues" about incorrect events can be misleading. For instance, \begin{quote}
      
  \textit{Furthermore, \textbf{expression} of an activated PKD1 mutant enhances HPK1-mediated NFkappaB activation.} \end{quote}
  Our model predicts \textit{\textbf{expression}} as a trigger of type \textit{Gene expression}, while the gold label is \textit{Positive regulation}. Despite that fact that our model is capable of handling such scenarios sometimes given grounded biomedical concepts and factual reasoning paths, there is still room for improvement. 
\end{itemize}



\section{Related Works}
\paragraph{Event Extraction} Most existing event extraction systems focus on extracting events in news. Early attempts relied on hand-crafted features and a pipeline architecture~\cite{gupta-ji-2009-predicting, li-etal-2013-joint}. Later studies gained significant improvement from neural architectures, such as convolutional neural networks \cite{chen-etal-2015-event, nguyen-grishman-2015-event}, and recurrent neural networks \cite{nguyen-etal-2016-joint}. More recent studies leverages large pre-trained language models to obtain richer contextual information \cite{wadden-etal-2019-entity, lin-etal-2020-joint}. Another line of works utilized GNN to enhance event extraction performance. \newcite{liu-etal-2018-jointly} applied attention-based graph convolution networks on dependency parsing trees. We instead propose a GNN, \GAENet, for integrating domain knowledge into contextualized embeddings from pre-trained language models.

\paragraph{Biomedial Event Extraction}
Event extraction for biomedicine is more challenging due to higher demand for domain knowledge. BioNLP 11 GENIA event extraction task (\GE) is the major benchmark for measuring the quality of biomedical event extraction system~\cite{kim2011overview}. Similar to event extraction in news domain, initial studies tackle biomedical event extraction with human-engineered features and pipeline approaches \cite{miwa2012boosting, bjorne-salakoski-2011-generalizing}. Great portion of recent works observed significant gains from neural models \cite{venugopal-etal-2014-relieving, rao-etal-2017-biomedical, jagannatha-yu-2016-bidirectional, bjorne-salakoski-2018-biomedical}. \newcite{li2019biomedical} incorporated information from Gene Ontology, a biomedical knowledge base, into tree-LSTM models with distributional representations. Instead, our strategy is to model two knowledge graphs from UMLS hierarchically with conceptual and semantic reasoning paths, providing stronger clues for identifying challenging events in biomedical corpus.
\section{Conclusion}
We have proposed a framework to incorporate domain knowledge for biomedical event extraction. Evaluation results on \GE{} demonstrated the efficacy of \GAENet{} and hierarchical KG representation in improving extraction of non-indicative trigger words associated nested events. We also show that our method is robust when applied to different amount of training data, while being advantageous in low-resource scenarios.  Future works include grounding adjective triggers to knowledge bases, better biomedical knowledge representation and extracting biomedical events at document level.

\section*{Acknowledgements}
We thank Rujun Han for helpful advice during the development of our model. We also appreciate insightful feedback from PLUSLab members, and the anonymous reviewers. This research was sponsored by an NIH R01 (LM012592) and the Intelligence Advanced Research Projects Activity (IARPA), via Contract No. 2019-19051600007. The views and conclusions of this paper are those of the authors and do not reflect the official policy or position of NIH, IARPA, or the US government.

\bibliographystyle{acl_natbib}
\bibliography{emnlp2020}

\clearpage
\appendix
\appendix

\section{Implementation Details}

Our models are implemented in PyTorch \cite{paszke2019pytorch}. Hyper-parameters are found by grid search within search range listed in Table \ref{table:hyperparameter}. The hyper-parameters of the best performing model are summarized in \ref{table:hyperparameter_best}. All experiments are conducted on a 12-CPU machine running CentOS Linux 7 (Core) and NVIDIA RTX 2080 with CUDA 10.1. 

To pre-train KGE, we leverage the TransE implementation from OpenKE \cite{han2018openke}. All tuples associated with the selected nodes described in Section \ref{sec:hierarchical_kg} are used for pre-training with margin loss and negative sampling,

{
\small
\begin{equation*}
  \mathcal{L} = \sum_{(h,\ell,t) \in S}\sum_{(h',\ell,t') \notin S}max(0, d(h,\ell,t)- d(h',\ell,t')+\gamma) 
\end{equation*}
}
where $\gamma$ denotes margin, and $d(x,x')$ denotes the $\ell-1$ distance between $x$ and $x'$. $h$ and $t$ are embeddings of head and tail entities from the gold training sets $S$ with relation $\ell$.  ($h'$,  $\ell$ ,$t'$) denotes a corrupted tuplet with either the head or tail entity replaced by a random entity. TransE is optimized using Adam \cite{kingma:adam} with hyper-parameters illustrated in Table \ref{table:kge_hyper_parameter}. Every 50 epochs, the model checkpoint is saved if the mean reciprocal rank on the development set improve from the last checkpoint; otherwise, training will be stopped.

\section{Dataset}
The statistics of \GE is shown in \ref{table:ge11_stats}. The corpus contains 14496 events with 37.2\% containing nested structure \cite{bjorne-salakoski-2011-generalizing}.\footnote{The dataset can be downloaded from \href{http://bionlp-st.dbcls.jp/GE/2011/downloads/}{http://bionlp-st.dbcls.jp/GE/2011/downloads/}.} We use the official dataset split for all the results reported.
\begin{table}[h] 
\small
\centering
\begin{tabular}{lc}
\hline 
\textbf{Hyper-parameter}  & \textbf{Range} \\ \hline
Relation MLP dim. & \{$300, 500, 700, 1000$\} \\
Trigger MLP dim. & \{$300, 500, 700, 1000$\} \\
Learning rate & \{ $1 \times 10^{-5}, 3 \times 10^{-5}$, $5 \times 10^{-5}$ \} \\

\hline
\end{tabular}
\caption{\label{font-table} Hyper-paramter search range for fine-tuning SciBERT.}
\label{table:hyperparameter}
\end{table}

\begin{table}[t]
\small
\centering
\begin{tabular}{lc}
\hline 
\textbf{Hyper-parameter}  & \textbf{Value} \\ \hline
Relation MLP dim. & $300$ \\
Trigger MLP dim. & $300$ \\
Learning rate &  $3 \times 10^{-5}$ \\
\GAENet{} node dim. & $300$ \\
\GAENet{} edge dim. & $300$ \\
\GAENet{} layers & 2 \\
Dropout rate & $0.2$ \\

\hline
\end{tabular}
\caption{\label{font-table} Hyper-paramters of the best performing \GAENet-SciBERT model.}
\label{table:hyperparameter_best}
\end{table}

\begin{table}[t]
\small
\centering
\begin{tabular}{lc}
\hline 
\textbf{Hyper-parameter}  & \textbf{Value} \\ \hline

Learning rate &  $0.5$ \\
Margin & $3$\\
Batch size & $128$ \\
\# corrupted tuplets / \# gold tuplets & $25$\\
\# Epochs & $500$ \\

\hline
\end{tabular}
\caption{\label{font-table} Hyper-paramters for pre-training KGE.}
\label{table:kge_hyper_parameter}
\end{table}

\begin{table}
\small
\centering
\begin{tabular}{lc}
\hline 
\textbf{Metric} & \textbf{Number} \\ \hline
events & $14496$ \\
sentences & $11581$ \\
nested events &  $37.2\%$ \\
intersentence events & $6.0\%$ \\

\hline
\end{tabular}
\caption{\label{font-table} \GE dataset statistics}
\label{table:ge11_stats}
\end{table}

\end{document}